\documentclass[letterpaper]{article} 
\usepackage{aaai2027}  
\usepackage[hyphens]{url}  
\usepackage{graphicx} 
\usepackage{natbib}  
\usepackage{caption} 
\usepackage{algorithm}
\usepackage{algorithmic}
\usepackage{amsmath}
\usepackage{amssymb}
\usepackage{newfloat}
\usepackage{listings}
\DeclareCaptionStyle{ruled}{labelfont=normalfont,labelsep=colon,strut=off} 
\floatstyle{ruled}
\newfloat{listing}{tb}{lst}{}
\floatname{listing}{Listing}

\usepackage{booktabs}

\def\our{COGENT}

\nocopyright 

\title{\our{}: Counterfactual Gaussian Explanations for Volumetric Medical Images}
\author{
    Dorian Rz\k{a}sa\textsuperscript{\rm 1},
    Bartosz Zabdyr \textsuperscript{\rm 1},
    Krzysztof Piekarz \textsuperscript{\rm 1},
    Jakub Grzywaczewski\textsuperscript{\rm 2},\\
    Bartlomiej Sobieski\textsuperscript{\rm 2, 3},
    Przemys{\l}aw Biecek\textsuperscript{\rm 2, 3},
    \.Zaneta \'Swiderska-Chadaj\textsuperscript{\rm 4},
    Olga \'Sliwicka\textsuperscript{\rm 5},\\
    Przemys{\l}aw Spurek\textsuperscript{\rm 1, 4},
    Joanna \'Swiebocka-Wi\k{e}k\textsuperscript{\rm 1}
}
\affiliations{
    \textsuperscript{\rm 1} Jagiellonian University\\
    \textsuperscript{\rm 2} Warsaw University of Technology\\
    \textsuperscript{\rm 3} University of Warsaw\\
    \textsuperscript{\rm 4} IDEAS Research Institute \\
    \textsuperscript{\rm 5} Medical University of Lodz
}

\begin{document}

\maketitle

\begin{abstract}
Explainability is essential for deploying deep learning models in high-stakes medical applications. Existing explainability methods for volumetric imaging predominantly operate in voxel space, overlooking the structured representations introduced by recent advances in 3D scene modeling. We present COGENT (Counterfactual Gaussian Explanations), a framework that generates counterfactual explanations directly in the parameter space of Gaussian-based volumetric representations. Built upon MedGS and the Sybil lung cancer risk prediction model, COGENT optimizes selected Gaussian primitives through a differentiable rendering pipeline, enabling gradients from the downstream predictor to identify representation components that most influence model decisions. Unlike conventional pixel- or voxel-level attribution methods, our approach formulates explainability as a counterfactual optimization problem over an explicit 3D scene representation, producing sparse and spatially localized explanations while preserving anatomical consistency. We evaluate COGENT on lung CT scans using quantitative comparisons with existing explainability methods together with qualitative analysis by medical experts. The results demonstrate that representation-space counterfactual optimization provides clinically meaningful explanations while offering a new perspective on interpreting volumetric deep learning models.
 Our code is publicly available at \url{https://github.com/gmum/COGENT}
\end{abstract}
\section{Introduction}

Explainable Artificial Intelligence (XAI)~\cite{xu2019explainable} has become a key requirement for the safe deployment of machine learning systems in high-stakes domains such as healthcare \cite{Holzinger2022,BANIECKI2025103026}. While modern deep learning models often achieve state-of-the-art predictive performance, their decision-making processes frequently remain opaque to end users. This lack of transparency is particularly problematic in medical imaging, where model predictions may influence diagnostic decisions and patient management \cite{ahmed2026explainable,rudin2019stop}.

\begin{figure}
    \centering
    \includegraphics[width=0.48\textwidth]{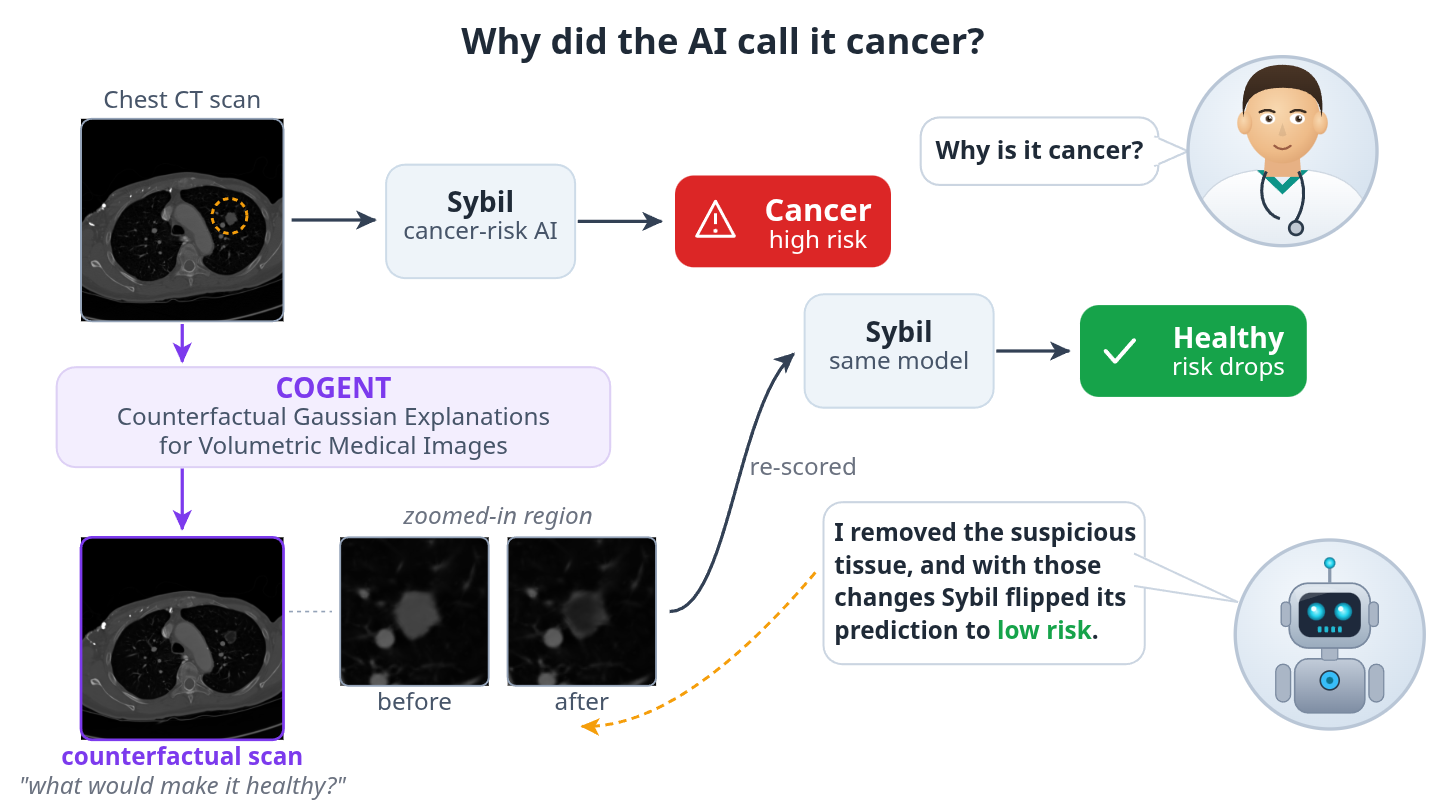}
    \caption{\small Overview of the COGENT framework. Given a scan that Sybil flags as high risk, COGENT asks what would make it healthy, removing the suspicious nodule with a localized Gaussian perturbation while leaving the surrounding anatomy untouched (before/after insets). Re-scoring the edited volume with the same model flips the prediction to low risk, revealing the evidence behind the original decision.}
    \label{fig:teasermaly}
\end{figure}

Recent years have seen rapid progress in volumetric deep learning systems operating directly on three-dimensional medical data \cite{mikhael2023sybil,marzol2025medgs}. Computed tomography (CT) and magnetic resonance imaging (MRI) provide rich spatial information that can be exploited by deep neural networks for classification, segmentation, risk prediction, and disease progression modeling \cite{wasserthal2023totalsegmentator,mikhael2023sybil,marzol2025medgs}. However, despite the inherently three-dimensional nature of these data, many explainability techniques continue to operate primarily on two-dimensional image representations \cite{ahmed2026explainable}. As a result, complex spatial dependencies may be obscured, making it difficult to understand how volumetric structures contribute to model predictions \cite{sobieski2026auditing}.

Lung cancer risk prediction represents a particularly important application of volumetric medical AI. Sybil is a deep learning model designed to estimate future lung cancer risk directly from a single low-dose chest CT scan \cite{mikhael2023sybil}. The model has demonstrated strong predictive capabilities and has become an important example of modern volumetric risk prediction systems. At the same time, its complexity motivates the development of interpretation techniques that provide clinically meaningful explanations of its predictions.

Existing approaches to model interpretation include saliency methods, gradient-based visualizations, feature attribution techniques, and counterfactual explanations \cite{selvaraju2017gradcam,lundberg2017unified,wachter2017counterfactual,zaher2024manifold}. While these methods have proven useful in many applications, they are typically formulated in image space and therefore operate on pixels or voxels. Such representations do not explicitly model the underlying three-dimensional structure of the scene and may produce explanations that are difficult to interpret anatomically \cite{chrabaszcz2025aggregated,ahmed2026explainable}.

Recent advances in 3D Gaussian Splatting (3DGS)~\cite{kerbl2023gaussian} have introduced explicit, differentiable scene representations that model complex volumetric structures using collections of Gaussian primitives. Unlike conventional image-based representations, Gaussian Splatting provides direct access to interpretable scene parameters, including position, appearance, opacity, scale, and orientation. Furthermore, MedGS~\cite{marzol2025medgs} extends this paradigm to multi-modal medical imaging and enables Gaussian-based representations of volumetric medical data.

In this work, we introduce \our{} (Counterfactual Gaussian Explanations for Volumetric Medical Images), a framework that directly embeds explainability within the three-dimensional representation space, see Fig.~\ref{fig:teasermaly}. Instead of perturbing raw image pixels or voxels, \our{} formulates interpretability as a constrained counterfactual optimization problem within the parameter space of an explicit 3D Gaussian representation. Specifically, our framework leverages MedGS to reconstruct the volumetric medical data, establishing a fully differentiable bridge between individual Gaussian primitives and rendered imaging slices. We then apply Projected Gradient Descent (PGD) to optimize selected Gaussian parameters, utilizing gradients propagated back from the frozen downstream Sybil classifier to actively shift the model's risk assessment. By combining this differentiable rendering pipeline with gradient-based optimization, the proposed approach enables the generation of multi-view consistent volumetric modifications. Ultimately, by analyzing and isolating the Gaussian primitives that undergo the most significant parameter updates, \our{} produces sparse, anatomically localized explanations that reveal the precise structural and textural regions driving the deep learning model's predictions.

The main contributions of this work are as follows.
\begin{itemize}
\item We introduce a representation-space explainability framework that connects MedGS representations with a volumetric lung cancer risk prediction model.
\item We formulate explainability as a counterfactual optimization problem directly in the parameter space of 3D Gaussian representations.
\item We demonstrate that Gaussian-space counterfactual explanations produce spatially localized and clinically meaningful explanations through quantitative evaluation and expert assessment.
\end{itemize}
\section{Related Works}

\our{} lies at the intersection of explainable artificial intelligence, adversarial and counterfactual explanations, medical risk prediction, and explicit three-dimensional scene representations. This section reviews previous work in these areas and positions the proposed framework with respect to existing approaches to interpret volumetric medical AI systems.

\paragraph{Explainable AI in Medical Imaging}
Explainable Artificial Intelligence has become an important research direction in medical imaging, where model predictions must be interpretable, reliable, and clinically meaningful \cite{molnar2024interpretable,samek2019explainable}. Classical post-hoc explanation methods aim to assign importance scores to input regions or features. Gradient-based saliency methods, Layer-wise Relevance Propagation, Integrated Gradients, Grad-CAM, and SHAP have been widely used to explain deep neural networks in image classification and medical imaging tasks \cite{bach2015lrp,sundararajan2017axiomatic,selvaraju2017gradcam,lundberg2017unified}.

\begin{figure*}[ht]
    \centering
    \includegraphics[height=0.3\paperheight,width=\textwidth]{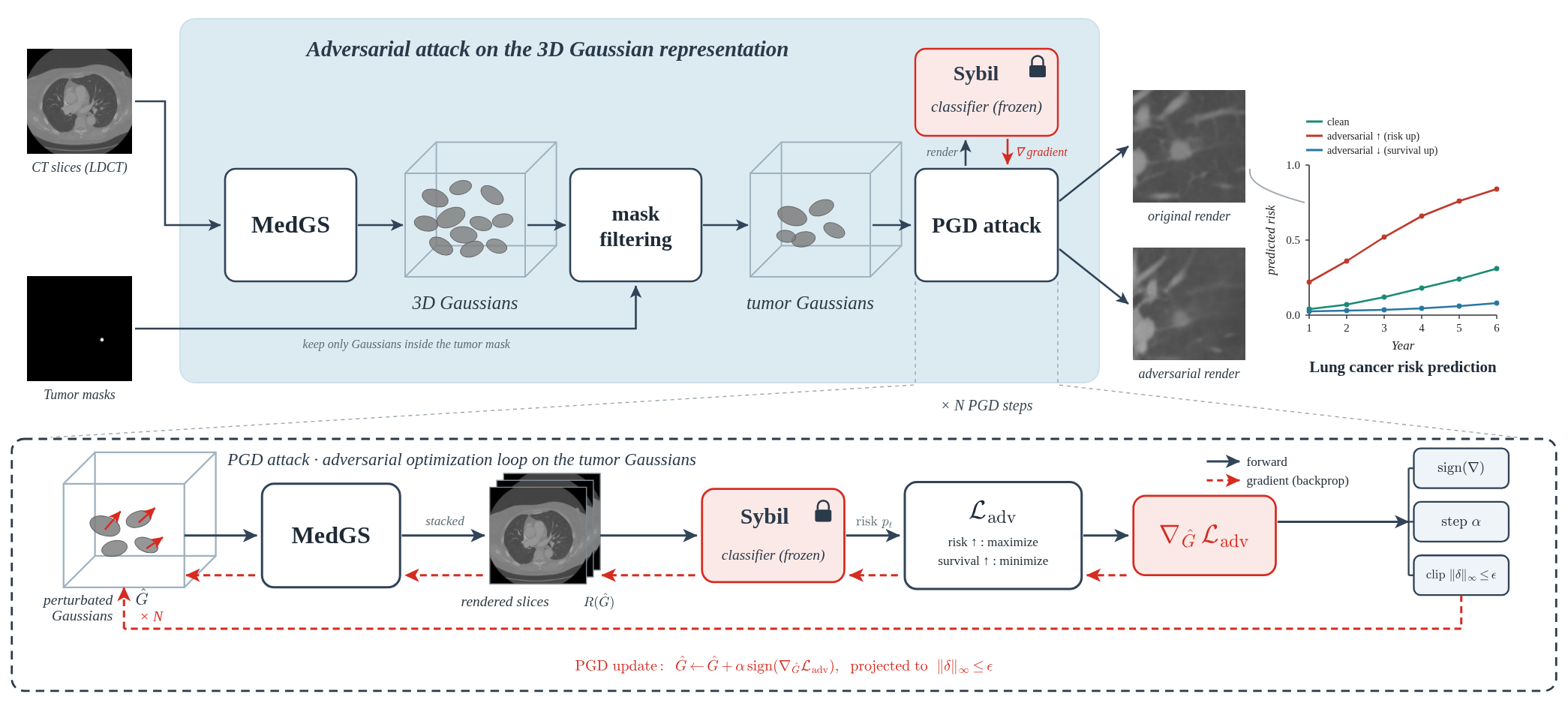}
    \caption{Gaussian-space counterfactual explanation generated for Sybil. An LDCT volume and its tumour mask are encoded by MedGS into 3D Gaussians, and mask filtering keeps only the Gaussians inside the tumour. Their parameters are optimized with PGD using gradients from the frozen Sybil classifier, under an $l_\infty$-$\epsilon$ bound, for $N$ steps. The resulting counterfactual render stays close to the original while shifting Sybil’s predicted 1–6-year risk upward or downward relative to the original prediction.}
    \label{fig:teaser}
\end{figure*}

For volumetric medical data, this creates an important limitation. CT and MRI scans contain three-dimensional anatomical structures, while many explanation methods are either applied slice-by-slice or produce heatmaps that may be difficult to interpret as coherent 3D objects. This motivates explanation strategies that better preserve spatial structure and operate on representations closer to the underlying anatomy.

\paragraph{Counterfactual and Adversarial Explanations}
Counterfactual explanations provide an alternative perspective on interpretability by asking how an input would need to change in order to alter a model prediction \cite{wachter2017counterfactual}. In computer vision, counterfactual methods move beyond highlighting important regions and instead attempt to generate modified inputs that reveal model-sensitive factors.

Adversarial methods provide another mechanism for probing model sensitivity. Originally introduced to study the vulnerability of neural networks to small perturbations \cite{goodfellow2015adversarial}, adversarial optimization and Projected Gradient Descent have become standard tools for analyzing and stress-testing deep models \cite{madry2018towards}. In the context of explainability, adversarial optimization can also be viewed as a mechanism for identifying components of the input representation that strongly influence model outputs.

\paragraph{Interpreting Lung Cancer Risk Models}
Sybil is a deep learning model for predicting future lung cancer risk from a single low-dose CT scan \cite{mikhael2023sybil}. Because it operates directly on volumetric CT data and predicts risk over a multi-year horizon, it represents an important example of a modern medical AI system that requires careful interpretability analysis.

Recent work on Sybil's audit proposed interventional explanation methods based on generative modifications of pulmonary nodules \cite{sobieski2026auditing}. In that framework, synthetic nodule removal and insertion are used to analyze the contribution of individual nodules and their interactions to model predictions. These results demonstrate the value of intervention-based approaches for understanding complex medical AI systems.

\our{} is complementary to this direction. Instead of performing interventions directly in the image space or in explicitly identified nodules, \our{} studies model behavior through an intermediate explicit three-dimensional representation.

\paragraph{Gaussian Splatting and Medical Gaussian Splatting}
3D Gaussian Splatting represents a scene as a collection of Gaussian primitives with parameters such as position, appearance, opacity, scale, and rotation \cite{kerbl2023gaussian}. The representation combines explicit scene structure with efficient differentiable rendering and provides direct access to interpretable scene components.

MedGS extends Gaussian Splatting to multi-modal three-dimensional medical imaging \cite{marzol2025medgs}. By combining Gaussian representations with medical volumes and differentiable rendering, MedGS provides a natural foundation for representation-level explainability methods. \our{} builds upon this idea and investigates whether model interpretation can be performed directly in Gaussian parameter space.
Unlike existing explainability methods that operate directly in image space, \our{} performs attribution and counterfactual optimization in an explicit three-dimensional Gaussian representation. To our knowledge, this is the first framework that generates Gaussian-space counterfactual explanations for volumetric lung cancer risk prediction.
\section{Preliminary studies}
\our{} builds upon Sybil, Gaussian Splatting, and MedGS. Below we briefly summarize these components. The overall optimization pipeline of the proposed \our{} framework is illustrated in Figure \ref{fig:teaser}.

\paragraph{Sybil}
Sybil is a deep learning framework developed for future lung cancer risk prediction from low-dose chest computed tomography (LDCT) scans \cite{mikhael2023sybil}. Unlike traditional risk assessment approaches, Sybil does not require manually annotated lesions, radiological findings, demographic information, or clinical risk factors during inference. Instead, the model estimates future lung cancer risk directly from imaging data.

In the proposed \our{} framework, Sybil is used as a fixed risk prediction model. Its output serves as the optimization signal that guides modifications of the underlying Gaussian representation.

\begin{figure}[h]
    \centering
    Original \qquad\qquad Baseline \qquad\quad \our{} (our) \\
    \includegraphics[width=0.45\textwidth,trim=80 140 500 83,
        clip]{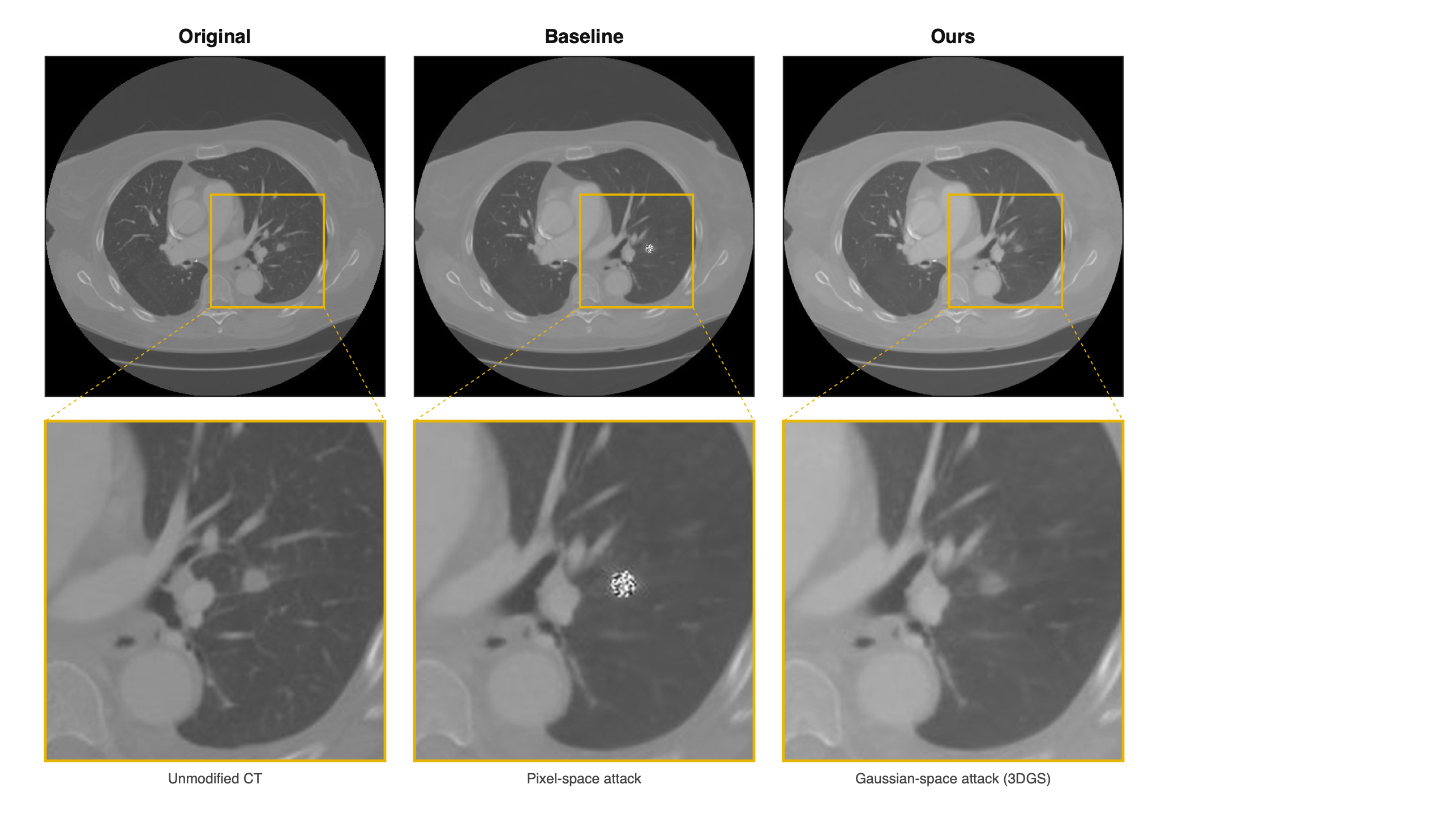}
    \caption{Qualitative comparison between the original CT slice, a pixel-space counterfactual baseline, and the proposed Gaussian-space explanation. Unlike the noisy pixel-space baseline, COGENT produces localized anatomically consistent counterfactual changes that reduce the predicted risk.}
    \label{fig:compare}
\end{figure}

\paragraph{Gaussian Splatting}

Gaussian Splatting is an explicit scene representation framework in which a scene is modeled as a collection of Gaussian primitives \cite{kerbl2023gaussian}. Each Gaussian is described by parameters including spatial position, appearance coefficients, opacity, scale, and rotation. The resulting representation can be rendered using a differentiable pipeline, allowing gradients to propagate from image-space objectives back to Gaussian parameters.

\begin{figure*}[ht]
    \centering
    \includegraphics[height =0.3\paperheight,width=\textwidth]{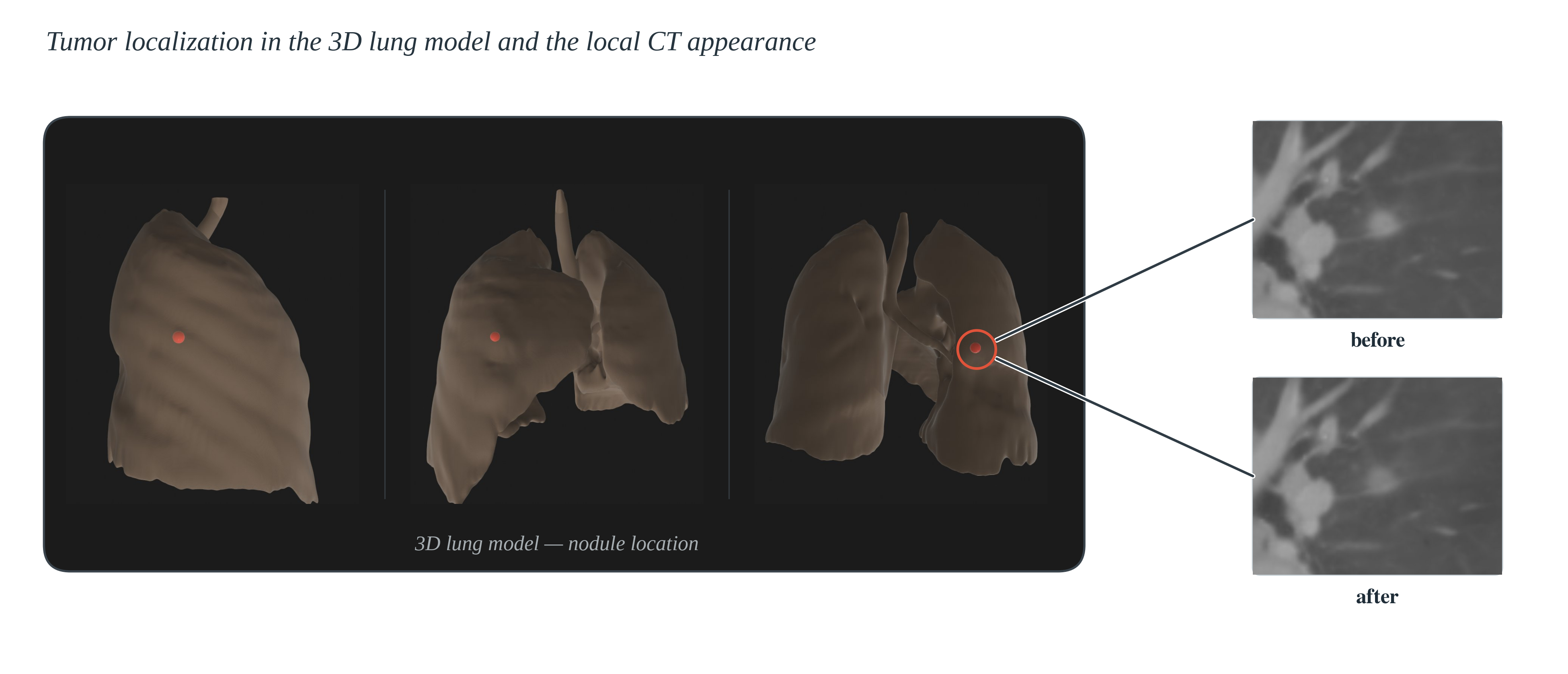}
    \caption{Tumour localisation in the 3D lung model and the corresponding local CT appearance. The highlighted Gaussian cluster identifies a lesion-bearing region, while the before/after crops show that the modification remains local and anatomically consistent.}
    \label{fig:lung}
\end{figure*}

The \our{}  framework leverages this explicit representation
to move explainability from image space to a structured
three-dimensional parameter space.

\paragraph{MedGS}

MedGS extends Gaussian Splatting to multi-modal three-dimensional medical imaging \cite{marzol2025medgs}. The framework represents medical volumes using Gaussian primitives and provides a differentiable rendering pipeline that enables reconstruction and analysis of volumetric medical data.

Within \our{}, MedGS serves as the intermediate representation layer between raw CT data and the Sybil. A medical volume is first represented using Gaussian primitives and subsequently rendered into slices that are assembled into a Sybil-compatible input volume. Because the entire pipeline remains differentiable, gradients originating from the Sybil can be propagated back to the Gaussian representation and used for explainability-driven optimization.

This property underpins the proposed \our{} framework and enables the generation of counterfactual explanations directly in Gaussian parameter space.


\section{Method description}
The key idea behind \our{} is to identify the Gaussian primitives that most strongly influence the prediction of a downstream medical AI model. Existing adversarial explanation methods typically operate in image space by perturbing pixels or voxels. In contrast, \our{} formulates explainability as an optimization problem in the parameter space of an explicit three-dimensional Gaussian representation. By optimizing selected Gaussian parameters through a fully differentiable rendering pipeline, the resulting changes reveal which components of the representation are most responsible for the model output.

We consider a scene represented by a set of $N$ Gaussians
\[
\Theta=\{\theta_i\}_{i=1}^N,\qquad
\theta_i=(\mathbf{x}_i,\mathbf{f}_i,\mathbf{o}_i,\mathbf{s}_i,\mathbf{r}_i),
\]
where $\mathbf{x}_i\in\mathbb{R}^3$ denotes the position, $\mathbf{f}_i$ the SH color coefficients, $\mathbf{o}_i$ opacity, $\mathbf{s}_i$ scales, and $\mathbf{r}_i$ rotation parameters. The counterfactual optimization is performed not on image pixels, but directly on a selected subset of scene parameters, after first identifying the Gaussians intersecting regions defined by anatomical masks and optionally discarding the largest geometric objects. In the implementation, this corresponds to the stages of mask filtering and subset selection of Gaussians before the actual optimization.

Let $\Omega\subset\{1,\dots,N\}$ denote the set of Gaussians selected for counterfactual optimization. For each camera $c_j$ (where each camera corresponds to a distinct 2D projection view, analogous to an imaging slice or an acquisition angle in medical tomography), the differentiable MedGS renderer produces an image
\[
\mathbf{I}_j(\Theta)=\mathcal{R}(\Theta,c_j).
\]
The rendered slices are then assembled into the input volume
provided to the Sybil classifier,
\[
\mathbf{V}(\Theta)=\mathcal{P}\big(\mathbf{I}_{j_1}(\Theta),\dots,\mathbf{I}_{j_D}(\Theta)\big),
\]
where $\mathcal{P}$ denotes the operator of ordering, resizing, and optional cropping/padding along the depth axis. The entire pipeline is differentiable, allowing gradients to propagate
from the classifier logits back to the Gaussian parameters.

The Sybil classifier is represented as an ensemble $\{f_m\}_{m=1}^M$, and its output consists of raw hazard logits
\[
\mathbf{z}_m=f_m(\mathbf{V}(\Theta))\in\mathbb{R}^6.
\]
The ensemble aggregation is
\[
\bar{\mathbf{z}}=\frac{1}{M}\sum_{m=1}^M \mathbf{z}_m.
\]
If counterfactual optimization targets a specific year $k^*\in\{0,\dots,5\}$, then the objective scalar is
\[
s(\Theta)=\bar{z}_{k^*}.
\]
If the averaged variant over all six years is used, then
\[
s(\Theta)=\frac{1}{6}\sum_{k=0}^{5}\bar{z}_k.
\]
In practice, the counterfactual optimization uses raw logits instead of calibrated probabilities, because sigmoid and clipping operations lead to saturation and weakened gradients in regions of high prediction confidence.

The objective function depends on the counterfactual goal. In the untargeted variant, the goal is to increase the selected risk logit, which we write as
\[
\mathcal{L}_{\mathrm{cf}}(\Theta)=-s(\Theta).
\]
In the targeted variant, the goal is to match the logit to a prescribed value $t$, so
\[
\mathcal{L}_{\mathrm{cf}}(\Theta)=\big(s(\Theta)-t\big)^2.
\]
In both cases, the gradient is computed with respect to the Gaussian parameters through the full differentiable chain
\[
\Theta \rightarrow \mathcal{R}(\Theta,c_j) \rightarrow \mathbf{V}(\Theta) \rightarrow f_m(\mathbf{V}(\Theta)) \rightarrow \mathcal{L}_{\mathrm{cf}}.
\]

The optimization follows a constrained Projected Gradient
Descent (PGD) procedure. For each optimized component $\theta_i^{(p)}$ and for $i\in\Omega$, we perform the iteration
\[
\theta_{i,t+1}^{(p)}=
\Pi_{[\ell,u]\cap B_\infty(\theta_{i,0}^{(p)},\varepsilon)}
\left(
\theta_{i,t}^{(p)}-\alpha\,\mathrm{sign}\!\left(\nabla_{\theta_{i,t}^{(p)}}\mathcal{L}_{\mathrm{cf}}\right)
\right),
\]
whereas for $i\notin\Omega$ the parameters remain unchanged. The operator $\Pi$ denotes projection onto the $L_\infty$ ball of radius $\varepsilon$ around the initial value and onto the global clipping interval $[\ell,u]$. Before starting the loop, a random start is applied:
\[
\theta_{i,0}^{(p)} \leftarrow \theta_{i,0}^{(p)} + \xi,\qquad \xi\sim \mathcal{U}(-\varepsilon,\varepsilon),
\]
which increases robustness to local minima and aligns with standard PGD practice.

The overall optimization procedure can be summarized as follows. First, a subset of Gaussians is determined on the basis of the masks and the scene geometry. The initial values of the optimized parameters are stored, and a random start is performed within the admissible perturbation ball. In each iteration, the selected slices are rendered, the input volume for Sybil is constructed, the objective scalar is computed from the raw logits, and then the loss and its gradient with respect to the scene parameters are evaluated. Finally, a PGD step is taken, after which the perturbation is projected back to the admissible set. Additionally, for interpretability, the implementation marks the 10\% most strongly changed Gaussians by setting their optimized parameters to a fixed visualization value.

\section{Experiment description}

The experiment addresses the following localization (attribution) question: \emph{which regions of the scan drive Sybil’s risk prediction, and do they coincide with the ground-truth tumour}? Rather than deriving attribution directly from gradients, we generate a counterfactual explanation by optimizing the scan in the \emph{Gaussian} representation and then analyze which Gaussian primitives must change to alter the predicted risk. Because the Gaussian representation is low-dimensional
and spatially localized, the Gaussians undergoing the largest
modifications form a compact and naturally sparse explanation
of the image content driving the model’s prediction. In the final step, these identified regions are compared with ground-truth tumour annotations.

The experiment proceeds in three stages, applied to each scan:
\begin{enumerate}
    \item generation of a counterfactual explanation in the Gaussian representation that changes Sybil's predicted prognosis,
    \item identification of the Gaussian primitives that undergo the largest changes during counterfactual optimization,
    \item visual comparison of these regions with the ground-truth tumour.
\end{enumerate}
All three stages are described below.

\paragraph{Counterfactual optimization in the Gaussian representation} 
Each LDCT volume is represented not as raw voxels but as a set of 3D Gaussian primitives (a Gaussian-splatting--style representation), each primitive carrying a position, a covariance (its scale and orientation), and an amplitude. Decoding this set reconstructs the volume that is fed to Sybil, so the predicted risk is a differentiable function of the Gaussian parameters.

Starting from the Gaussians reconstructing the original scan,
counterfactual optimisation updates their parameters to generate
a counterfactual explanation that shifts Sybil’s predicted risk
in a desired direction: either \emph{increasing} the predicted risk (worsening the prognosis) or \emph{decreasing} it (improving the prognosis). The optimization is constrained so that the perturbed volume remains close to the original, yielding only minor image changes while producing a substantial change in the predicted risk. The output of this stage is a perturbed set of Gaussians for each scan, together with the corresponding change in the predicted risk.

\paragraph{Selecting the most-changed Gaussians}
Given the original and counterfactually modified Gaussians, each primitive is scored according to the magnitude of the changes introduced during the counterfactual optimization (for example, the displacement of its position and the change in its amplitude or scale). The Gaussians are ranked by this change score and the most strongly altered ones are retained (the top-ranked subset). By construction, these are the primitives whose modification
most strongly influences Sybil's prediction and therefore
identify the regions to which the model is most sensitive for
that scan.

\paragraph{Expert Qualitative Evaluation of Counterfactual Explanations}
To evaluate whether COGENT highlights clinically meaningful imaging features, we establish an expert qualitative evaluation protocol based on Gaussian-space counterfactual explanations. The selected Gaussians are projected back to their corresponding 3D CT locations and overlaid on the original scans with ground-truth tumour annotations (confirmed via radiologist labels or follow-up imaging). An experienced radiologist visually inspects these overlays to evaluate whether peak model perturbations directly correspond to alterations in clinical disease severity. Grading follows a 5-point scale: positive ratings ($+2, +1$) denote attenuation of malignancy markers (e.g., reduction in tumour volume, structural complexity, or signal intensity); $0$ indicates a neutral impact; and negative ratings ($-1, -2$) denote exacerbation of features associated with disease progression.

The primary outcome of the experiment is therefore qualitative and spatial: the degree of correspondence, across scans, between the most strongly perturbed Gaussians and the ground-truth tumour location. A high correspondence supports the interpretation that the counterfactual explanation correctly localises the evidence underlying the model's prediction to the lesion. A low or inconsistent correspondence would instead indicate that Sybil's prediction is driven by features away from the visible tumour.

\begin{table*}[]
\centering
\caption{Quantitative comparison of explainability methods on the Sybil model. \our{} achieves the highest RRA and RRA\_abs values among the evaluated methods while preserving high sparsity and competitive perturbation-based metrics.}
\label{tab:Results}

\resizebox{0.8\textwidth}{!}{%
\begin{tabular}{lccccc}
\toprule
\textbf{Method} & \textbf{RRA} & \textbf{RRA\_abs} & \textbf{Perturbation\_AUC} & \textbf{Mean\_Perturbation} & \textbf{Sparseness} \\
\midrule
\textbf{\our{}} & \textbf{0.2837} & \textbf{0.2837} & \textbf{0.0396} & \textbf{0.0414} & \textbf{0.6550} \\
Saliency & 0.0702 & 0.0702 & 0.0375 & 0.0393 & 0.6182 \\
Input $\times$ Gradient & 0.0218 & 0.0218 & 0.0395 & 0.0414 & 0.6288 \\
Integrated Gradients & 0.0254 & 0.0254 & 0.0384 & 0.0403 & 0.6700 \\
GradCAM & 0.0805 & 0.0805 & 0.0263 & 0.0276 & 0.6568 \\
Kernel SHAP & 0.0134 & 0.0134 & 0.0293 & 0.0306 & 0.2083 \\
\bottomrule
\end{tabular}%
}
\end{table*}

\section{Results}

Qualitatively, the proposed Gaussian-space counterfactual generation produces perturbations that are substantially more anatomically plausible than the pixel-space baseline. As shown in Figure~\ref{fig:compare}, the baseline alters the CT crop in a scattered, speckle-like fashion, introducing a visually noisy pattern that does not correspond to a coherent anatomical structure. By contrast, the Gaussian-space counterfactual generation concentrates the modification in a compact region that overlaps with a lesion-like area. This is important because the counterfactual modification is not merely visible; it is localized to the region where an experienced reader would expect to find a tumour. In other words, the method modifies the image in regions that are most relevant to the model's prediction.

\begin{figure*}[!h]
    \centering
    \includegraphics[width=0.99\textwidth]{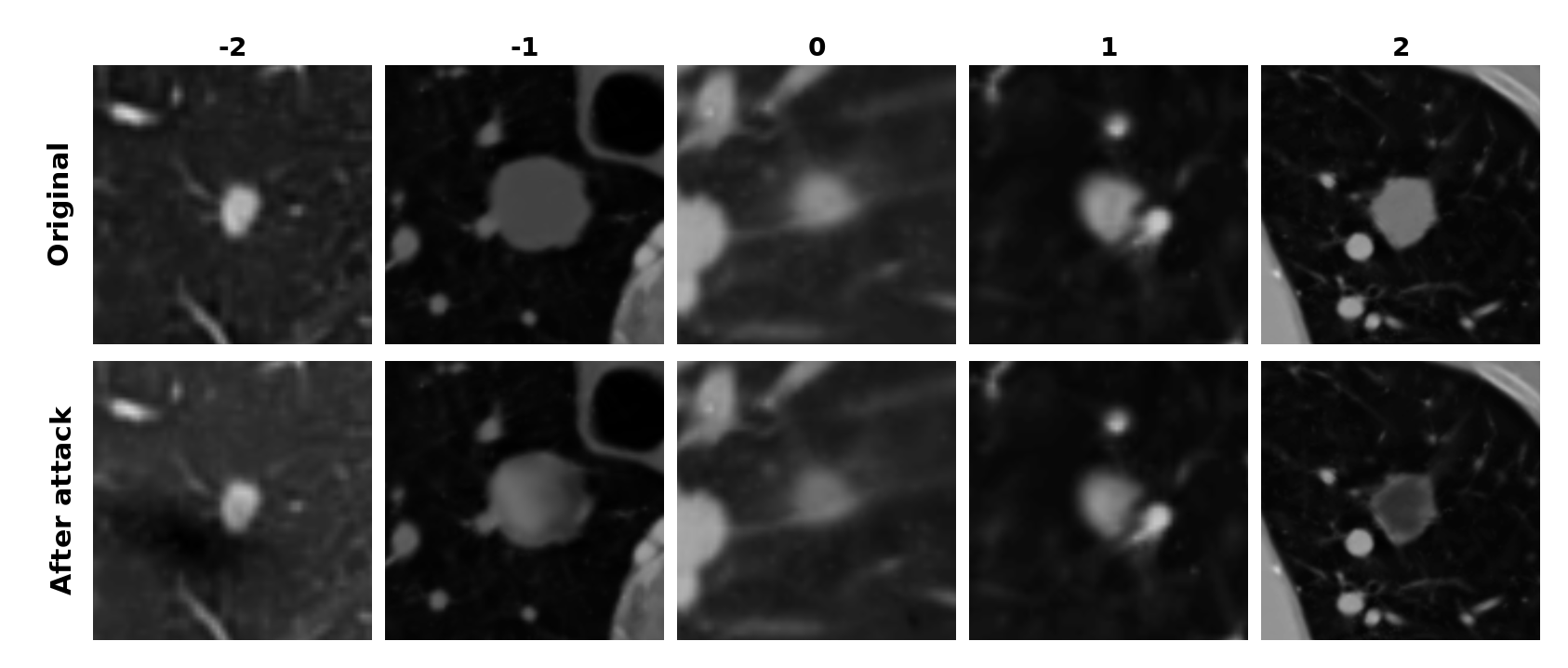}
    \caption{Representative counterfactual explanations generated by \our{}. Expert ratings are shown above each example.}
    \label{fig:before_after}
\end{figure*}

The same conclusion is reinforced by the 3D lung visualization shown in Figure~\ref{fig:lung}. The reconstructed lung model shows that the relevant Gaussian cluster is localized to a specific nodule-bearing region, and the corresponding before/after crops indicate that the counterfactual modification remains spatially local rather than diffusing across the entire lung field. This localization is not coincidental. It demonstrates that the model modifies precisely the region that an expert reader would identify as suspicious, which is exactly what is required for a clinically meaningful
counterfactual explanation. The before-and-after panels also show that the local CT appearance changes in a way consistent with the nodule location indicated in the 3D model.

The quantitative results in Table~\ref{tab:Results} confirm the visual findings. To ensure a fair comparison between methods operating in different representational spaces, all attribution maps, whether generated in pixel space (baselines) or Gaussian space (\our{}), are projected onto the same 2D slice plane and normalized to a common scale before evaluation. The localization score measures the overlap between the attribution map and the ground-truth lesion mask, thereby assessing how accurately each method identifies the pathological region. Specifically, we report two variants: RRA (Relative Random Area) and RRA\_abs, which quantify the fraction of positive attributions within the lesion relative to a random baseline, with the latter using absolute attribution values to account for sign inconsistencies between methods. \our{} achieves the best localization score, with RRA and RRA\_abs equal to 0.2837. This value is markedly higher than the strongest classical baselines: GradCAM (0.0805), Saliency (0.0702), Integrated Gradients (0.0254), Input $\times$ Gradient (0.0218), and Kernel SHAP (0.0134). These results demonstrate that operating directly in Gaussian space yields more spatially
concentrated and anatomically meaningful explanations than
pixel-space gradient- or perturbation-based methods.

At the same time, \our{} maintains a strong degree of compactness, with Sparseness equal to 0.6550. This places it among the most focused methods in the comparison and well above Kernel SHAP, which is much more diffuse at 0.2083. The Perturbation-AUC results are competitive as well: \our{} scores 0.0396, which is close to Saliency (0.0375), Input $\times$ Gradient (0.0395), and Integrated Gradients (0.0384), while remaining above GradCAM (0.0263) and Kernel SHAP (0.0293). Mean-Perturbation follows the same pattern, with \our{} at 0.0414. Taken together, these results indicate that \our{} achieves strong localization while maintaining competitive stability across all three evaluation criteria.

Results from the expert qualitative evaluation confirm that \our{}-induced perturbations directly correlate with clinically interpretable features. Representative examples are presented in Figure \ref{fig:before_after}. The figure illustrates counterfactual explanations spanning the full range of expert ratings. Expert assessment indicated positive mitigation of disease severity (ratings $+1/+2$) in $40\%$ of evaluated cases, characterized by attenuated malignant traits. A neutral impact (rating $0$) was observed in $40\%$ of cases, while $20\%$ exhibited an exacerbation of progression-associated features (ratings $-1/-2$). Overall, these findings demonstrate that Gaussian-space counterfactual explanations effectively localize radiologically relevant disease features. These findings are consistent with the clinical understanding that lung cancer prognosis depends not only on tumour morphology but also on anatomical location and the surrounding parenchymal context.

Overall, the experimental results support the central claim of the paper. The Gaussian-space formulation produces counterfactual explanations that are visually cleaner than the pixel-space baseline and, more importantly, localize to CT regions that an expert would reasonably consider tumour-like. Figure~\ref{fig:compare} shows the contrast between the chaotic baseline and the structured Gaussian-space perturbation; Figure~\ref{fig:lung} further demonstrates that the modification remains anatomically localized in 3D, and the table in Table~\ref{tab:Results} shows that the proposed method yields the strongest localization performance among the evaluated approaches. The result is a method that is not only technically effective but also easier to interpret and justify from a clinical perspective.

\section*{Conclusion}

We introduced \our{}, a framework for representation-space explainability in 3D medical AI that operates directly in the parameter space of 3D Gaussian representations. Applied to the Sybil 3D CT lung cancer risk prediction model, COGENT demonstrates through both quantitative metrics and expert physician evaluation that its counterfactual explanations consistently highlight clinically relevant radiological features and localized disease regions. Ultimately, COGENT demonstrates the potential of counterfactual parameter-space explanations as a general framework for transparent and clinically meaningful interpretation of volumetric AI models. More broadly, the proposed framework is applicable to a wide range of volumetric AI systems and provides a general paradigm for representation-space explainability beyond the specific Sybil application considered in this work.

\section{APPENDIX}
\paragraph{}
The experiment was conducted to analyze the sensitivity of the Sybil model's predictions to perturbations introduced by an iterative counterfactual optimization, where the model outputs a baseline survival probability $Pr(x)$ for an input image $x$. For five distinct values of the step-scaling parameter ($\varepsilon \in \{0.1, 0.5, 1.0, 2.0, 4.0\}$), the counterfactual optimization procedure was executed for exactly 10 update steps. To quantify the effect of the perturbations on the model's decision, we use the average percentage change per step:
\[
\Delta(\varepsilon) = \frac{\left| Pr(x) - Pr(x_{n}) \right|}{n},
\]
where $Pr(x_{n})$ denotes the survival probability after all $n$ counterfactual optimization steps for a given $\varepsilon$. The measured results are summarized below:

\begin{table}[htbp]
\centering
\begin{tabular}{cc}
\toprule
$\varepsilon$ & Average probability change per step [\%] \\
\midrule
0.1 & 0.380 \\
0.5 & 1.908 \\
1.0 & 3.056 \\
2.0 & 4.168 \\
4.0 & 4.772 \\
\bottomrule
\end{tabular}
\caption{Average change in probability per step for the COGENT counterfactual optimization}
\label{tab:cogent_eps}
\end{table}

\begin{figure}[htbp]
\centering
\includegraphics[width=1.0\linewidth]{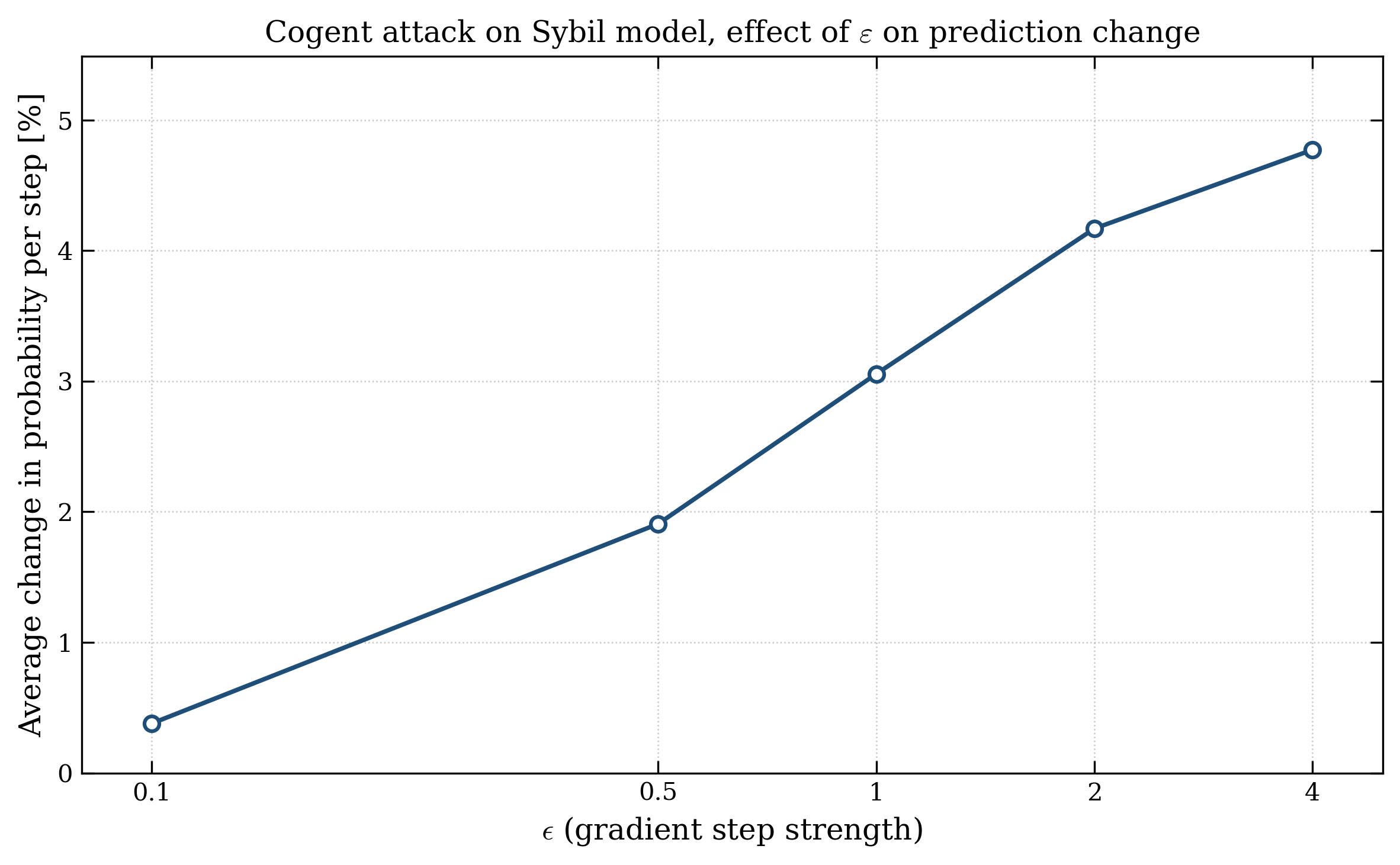}
\caption{Average change in probability per step as a function of $\varepsilon$ for the COGENT counterfactual optimization.}
\label{fig:cogent_eps}
\end{figure}

The results indicate a clear monotonic trend: increasing the perturbation magnitude $\varepsilon$ consistently raises the average per-step shift in the predicted probability. At the lowest setting ($\varepsilon = 0.1$), the model exhibits only a negligible response, with a change of approximately $0.38$ percentage points per step. This indicates that very small perturbations lie below the effective sensitivity threshold of the decision function and are largely absorbed by the model's robust feature representations. As $\varepsilon$ increases, the per-step shift grows accordingly, consistent with the monotonic relationship noted above.

\paragraph{}
In conclusion, the results suggest a strong coupling between the perturbation magnitude and the average response of the Sybil model. The highest relative increase in prediction volatility occurs for $\varepsilon$ values between 0.1 and 1.0. For larger perturbation scales, the effect gradually plateaus, suggesting that the model's output reaches a saturation zone where additional perturbations yield only marginal changes in the estimated survival risk. This saturation is also consistent with the fact that the predicted survival probability is inherently bounded within the $[0, 100]\%$ range, so the per-step shift cannot grow indefinitely and must diminish as the probability approaches its limits.

\begin{figure}[t]
\centering
\includegraphics[width=1.0\linewidth]{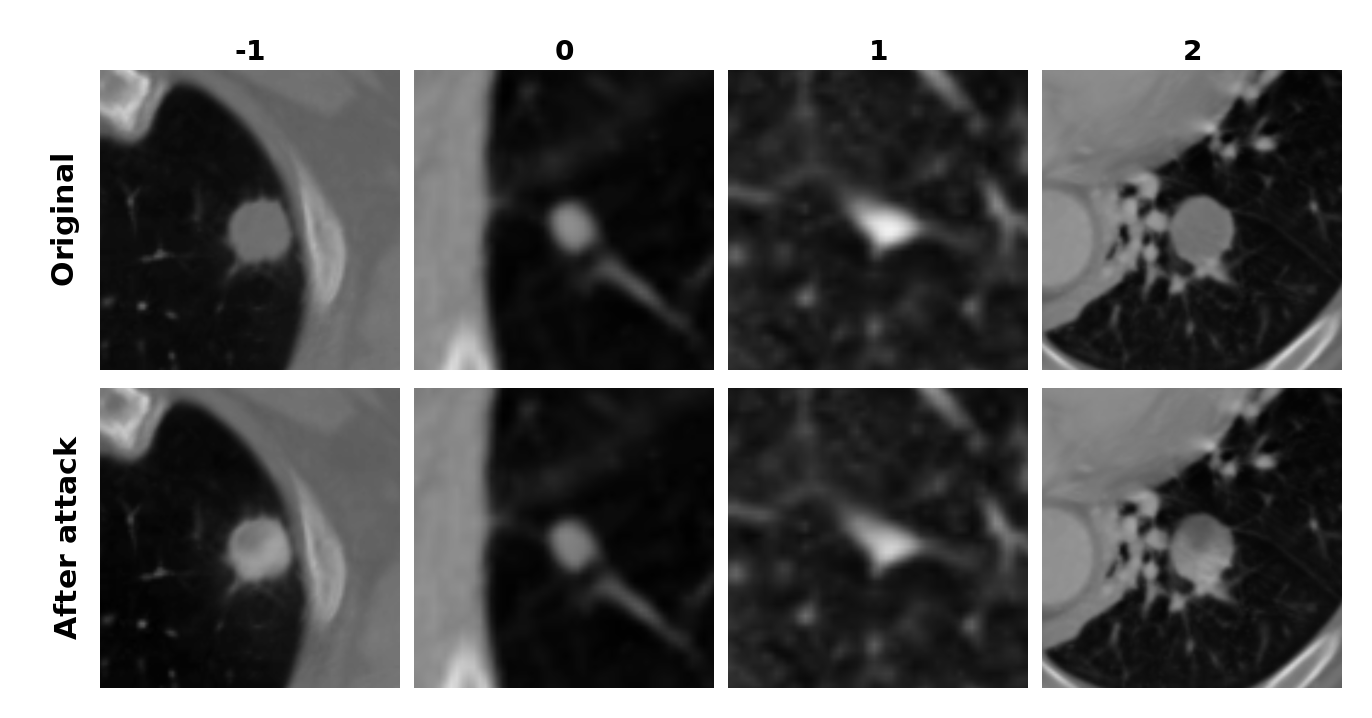}
\caption{Original axial CT slices (top) and their corresponding \our{} counterfactual explanations (bottom). Numbers denote radiologist annotations.}
\label{fig:cogent_qualitative}
\end{figure}

\bibliography{aaai2027}

\end{document}